\pdfoutput=1

\documentclass[11pt]{article}

\usepackage[preprint]{acl}
\usepackage[normalem]{ulem}
\useunder{\uline}{\ul}{}
\usepackage{comment}
\usepackage{times}
\usepackage{latexsym}
 \usepackage{multirow}
 \usepackage{booktabs}
 \usepackage{algorithm}
 \usepackage{algpseudocode}
 \usepackage{algorithmicx}

\usepackage[T1]{fontenc}
\usepackage{enumitem}

\usepackage[utf8]{inputenc}
\usepackage{amsmath} 
\usepackage{microtype}

\usepackage{inconsolata}
\usepackage{dsfont}
\usepackage{graphicx}

\title{Decoupling Reasoning and Knowledge Injection for \\ In-Context Knowledge Editing}

\usepackage{authblk}
\author[1]{\textbf{Changyue Wang}\thanks{cy-wang24@mails.tsinghua.edu.cn}}
\author[1]{\textbf{Weihang Su}}
\author[1]{\textbf{Qingyao Ai}\thanks{Corresponding Author: aiqy@tsinghua.edu.cn}}
\author[1]{\textbf{Yujia Zhou}}
\author[1]{\textbf{Yiqun Liu}}

\affil[1]{Department of Computer Science and Technology, Tsinghua University}

\begin{document}
\maketitle
\begin{abstract}

Knowledge editing aims to efficiently update Large Language Models (LLMs) by modifying specific knowledge without retraining the entire model. 
Among knowledge editing approaches, in-context editing (ICE) offers a lightweight solution by injecting new knowledge directly into the input context, leaving model parameters unchanged.
However, existing ICE approaches do not explicitly separate the newly injected knowledge from the model’s original reasoning process. 
This entanglement often results in conflicts between external updates and internal parametric knowledge, undermining the consistency and accuracy of the reasoning path.
In this work, we conduct preliminary experiments to examine how parametric knowledge influences reasoning path planning. We find that the model’s reasoning is tightly coupled with its internal knowledge, and that naively injecting new information without adapting the reasoning path often leads to performance degradation, particularly in multi-hop tasks.
To this end, we propose DecKER, a novel ICE framework that decouples reasoning from knowledge editing by generating a masked reasoning path and then resolving knowledge edits via hybrid retrieval and model-based validation.
Experiments on multi-hop QA benchmarks show that DecKER significantly outperforms existing ICE methods by mitigating knowledge conflicts and preserving reasoning consistency. \footnote{Our code is available at: https://github.com/bebr2/DecKER}

\end{abstract}

\section{Introduction}

Large language models (LLMs) have achieved remarkable performance across a wide range of knowledge-intensive tasks~\cite{openai2024gpt4technicalreport,fang2024scaling,deepseekai2025deepseekr1incentivizingreasoningcapability,qwen2.5}. 
However, their reliance on static parametric knowledge acquired during pre-training makes it difficult to keep them aligned with evolving real-world facts.
While continued pre-training offers a straightforward solution for updating model knowledge, it is computationally expensive and impractical for frequent updates.
To address this limitation, knowledge editing has emerged as a promising alternative. Instead of retraining the entire model, knowledge editing aims to efficiently modify specific factual associations within an LLM, allowing for low-cost, targeted updates that preserve the model’s overall capabilities.

Existing knowledge editing approaches can be broadly categorized into two categories: \textit{parameterized editing}, which updates the model’s weights to encode new facts (e.g., ROME \cite{meng2023locatingeditingfactualassociations}), and \textit{in-context editing} (ICE), which introduces new information into the model’s input without modifying parameters~\cite{wang2024knowledge,wang2024lekube}. 
While parameterized approaches offer persistent updates, they often suffer from side effects such as degraded performance on unrelated tasks~\cite{yang2024fall,gu24modeleditingharms, li2024reallyeditlanguagemodels} and struggle in complex reasoning scenarios~\cite{zhong2024mquakeassessingknowledgeediting}. 
In contrast, ICE methods aim to address the limitations of parameterized editing by preserving the original parameters of LLMs and directly editing the input context (e.g., prompts) to guide LLMs to behave based on the updated knowledge~\cite{cohen2023evaluatingrippleeffectsknowledge,wang2024knowledge,zhong2024mquakeassessingknowledgeediting}.

In principle, ICE methods should preserve the reasoning capabilities of LLMs by editing only the input context and avoiding parameter modifications. 
However, our empirical observations reveal a surprising contradiction: existing ICE methods often lead to substantial performance degradation. In multi-hop QA tasks, for instance, injecting new facts into the context can reduce accuracy by over 80\% (see \autoref{sec:preli}). This discrepancy challenges the core assumption that editing the context preserves the model’s reasoning behavior.

To understand this phenomenon, we conduct preliminary experiments to analyze how ICE influences the model’s reasoning behavior.
Our results indicate that when LLMs answer multi-hop questions based on their internal parametric knowledge, they demonstrate strong reasoning capabilities through high-quality chain-of-thought (CoT) reasoning path~\cite{wei2023chainofthoughtpromptingelicitsreasoning}. 
However, after new knowledge is injected into the input context, the model’s reasoning process is often affected. The conflict between the injected information and the model’s internal knowledge causes the LLM to stray from its original chain-of-thought reasoning, making it effectively “forget” how to solve the task.
This indicates that the reasoning ability of LLMs is closely tied to their internal knowledge, and adding conflicting context through ICE can interfere with the generation process and significantly degrade the model’s performance.
This phenomenon could limit the future applications of ICE methods, particularly when inference scaling and reasoning LLM (like Deepseek-R1) has recently become the dominant direction for super-intelligence development ~\cite{deepseekai2025deepseekr1incentivizingreasoningcapability}.

In light of these findings, we propose DecKER, a novel in-context editing method that \uline{Dec}ouples \uline{K}nowledge \uline{E}diting and model \uline{R}easoning. 
Taking multi-hop QA as an example, DecKER first extracts a masked reasoning path from LLMs, where entities subject to potential edits are replaced with placeholders along with corresponding type hints. 
Then, for each placeholder, we propose a hybrid mechanism combining retrieval-based conflict detection and model judgment to determine whether the entity is related to edited knowledge and fill it accordingly. 
Furthermore, DecKER samples multiple reasoning paths and selects the most consistent candidate using a set of scoring criteria, ensuring that the model retains its original reasoning structure while accurately integrating new facts.
Experimental results on multiple knowledge editing benchmark datasets show that DecKER significantly outperforms existing ICE methods by mitigating knowledge conflicts and preserving reasoning consistency.

In summary, this paper makes the following key contributions:
\begin{enumerate}[leftmargin=*]
\item We explore the impact of the conflict between injected contextual knowledge and LLMs' parametric knowledge, revealing how entangled reasoning and editing processes lead to significant performance degradation in multihop tasks.

\item We introduce DecKER, a novel ICE framework that employs global planning to decouple reasoning from knowledge injection via masked reasoning path generation.

\item We conduct comprehensive experiments to demonstrate that decoupling reasoning from knowledge injection significantly enhances multi-hop reasoning performance, addressing the limitations of prior ICE approaches.
\end{enumerate}





\section{Related Work}

Knowledge Editing (KE) aims to efficiently update knowledge in LLMs, divided into parameterized and non-parameterized methods \cite{wang2024lekube}. ROME \cite{meng2023locatingeditingfactualassociations}, a typical parameterized method, uses causal intervention to locate and edit related neurons, while MEMIT\cite{meng2023masseditingmemorytransformer} extends this capability to handle larger edit batch. Non-parameterized methods leverage the LLM's in-context learning\cite{brown2020languagemodelsfewshotlearners} abilities. Mello \cite{zhong2024mquakeassessingknowledgeediting} employs in-context editing (ICE) to tackle complex problems by breaking them into subtasks and performing fine-grained edits. Built on this, PokeMQA\cite{gu2024pokemqaprogrammableknowledgeediting} enhances robustness and DeepEdit\cite{wang2024deepedit} focuses on the reasoning process. Additionally, \citet{shi24rae} addresses multi-hop tasks in knowledge editing by emphasizing the retrieval process with their RAE method, which employs knowledge graph editing and retrieval to boost multi-hop reasoning performance.

To evaluate LLMs' reasoning abilities post-editing, \citet{cohen2023evaluatingrippleeffectsknowledge} introduces the concept of Ripple Effects, where altering one piece of knowledge can impact related facts. \citet{zhong2024mquakeassessingknowledgeediting} develops a multi-hop QA dataset to assess if edited models can utilize new knowledge for complex reasoning. Their findings indicate that ICE methods outperform parameterized methods in managing Ripple Effects and complex reasoning.

\section{Preliminary Study}
\label{sec:preli}
This section begins by introducing the multi-hop question-answering (MQA) task under knowledge editing. Then, we discuss the phenomenon of reasoning degradation in existing ICE methods and design an analytical experiment to explain it.

\subsection{MQA under Knowledge Editing}

Knowledge editing in the MQA task involves modifying the object component of a knowledge triple. Given an initial fact $e=(s,r,o)$, comprising a subject ($s$), a relation ($r$), and an object ($o$), this fact is edited to $e^* = (s, r, o^*)$. For a multi-hop question $Q$ and a set of edited facts $\mathcal{E}$ associated with $Q$, the reasoning path $P_{Q, \mathcal{E}}$ is represented as:

{\small \begin{equation}
    <e_1, \dots, e_n> = <(s_1, r_1, o_1), \dots, (s_n, r_n, o_n)>,
\end{equation}}

\noindent where $n$ is the number of reasoning hops, $s_{i+1} = o_i$, and $o_n$ is the final answer. The reasoning path before editing is denoted as $P_{Q, \emptyset}$.
If $\mathcal{E}$ is non-empty, indicating modifications to one or more knowledge triples in $Q$ (e.g., $e_k = (s_k, r_k, o_k)$ is edited into $(s_k, r_k, o_k^*)$), the reasoning path $P_{Q, \mathcal{E}}$ will reflect changes to $e_k$ and all subsequent triples, thus altering the final answer. 

Notably, successfully addressing MQA depends on two key components: accurate reasoning path planning and precise knowledge injection. Reasoning path planning involves determining the reasoning framework, i.e., the list of relations $R_Q = [r_1, \dots, r_n]$ in the reasoning path $P_{Q, \mathcal{E}}$, while knowledge injection ensures the model accurately provides $o_k$ given $(s_k, r_k)$. In real tasks, the edit batch size (the size of the union set of all $\mathcal{E}$) is often greater than one, and one reasoning path $P_{Q, \mathcal{E}}$ may involve multiple edits, adding significant complexity to knowledge editing methods.


\begin{figure}[t]
        \centering
        \includegraphics[width=\linewidth]{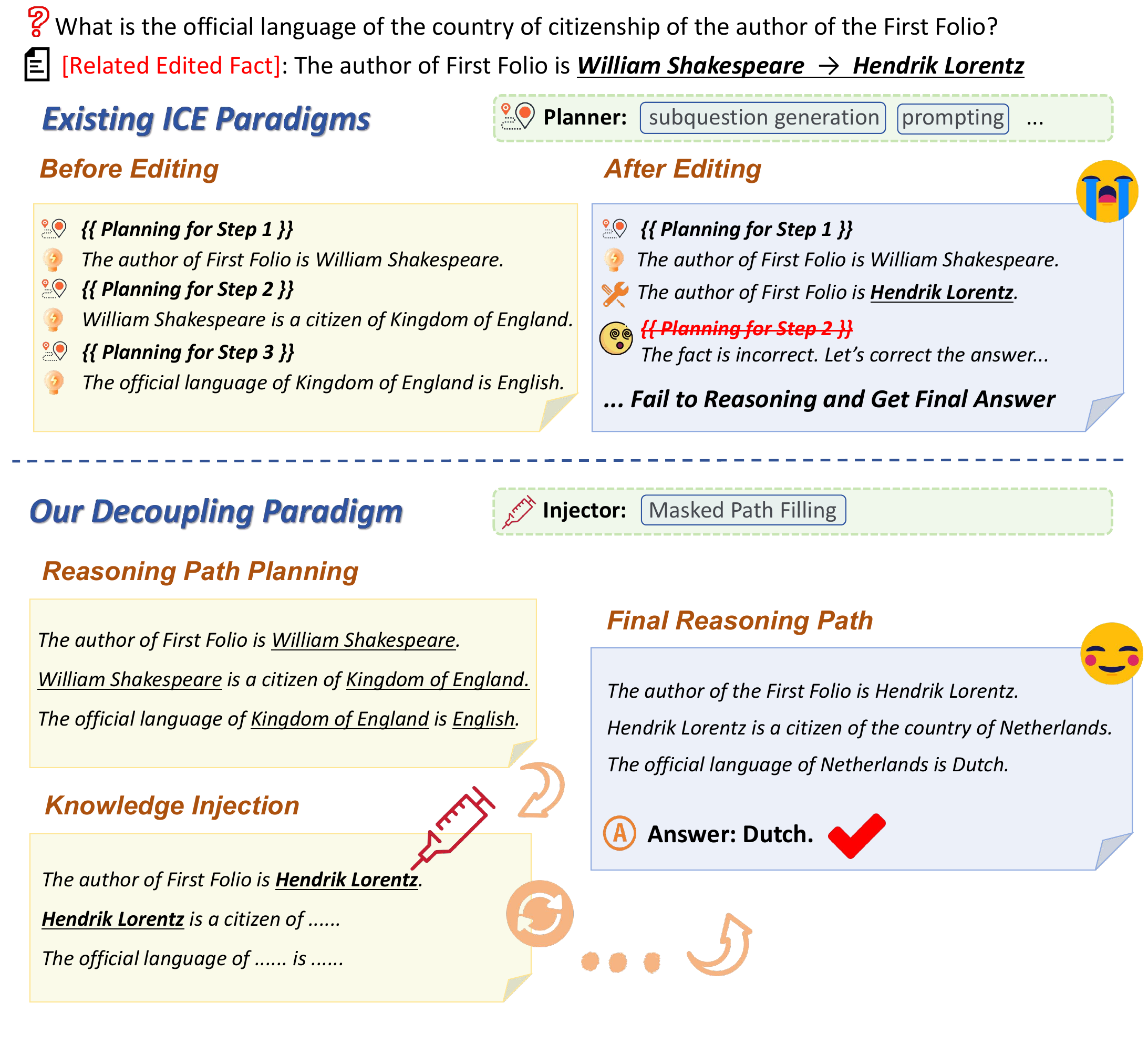}
        \caption{The comparison of existing ICE paradigms and our proposed decoupling paradigm. }
        \label{fig:analysis}
\end{figure}

\subsection{Reasoning Degradation in ICE}
\label{sec:sec2.2}

The upper part of \autoref{fig:analysis} illustrates a typical example of Reasoning Degradation in current ICE methods. These methods can be abstracted as interactions between a \textit{Planner}, which determines the next reasoning step through sub-question generation (e.g., Mello and PokeMQA) or prompting (e.g., DeepEdit and EditCoT), and an \textit{Editor}, which incorporates edited knowledge by replacing intermediate answers or prompting the model to revise them. Before the relevant knowledge is edited, the LLM uses its internal knowledge for reasoning, allowing the interaction between the Planner and Editor to proceed normally and produce the correct reasoning path. However, after the knowledge is edited and the Editor replaces the contextual knowledge with new one, the Planner does not continue reasoning as expected. Instead, it shifts direction, such as attempting to resolve conflicting knowledge. Moreover, as demonstrated in \autoref{apd:casestudy}, we present additional examples where the Planner influenced by edited knowledge exhibits behaviors like causing reasoning interruptions or redirecting to different reasoning paths. Clearly, current ICE methods fail to separate knowledge from reasoning, thereby impairing the reasoning abilities of LLMs. We design an experiment to quantitatively assess the Reasoning Degradation phenomenon.

\begin{figure}[t]
        \centering
        \includegraphics[width=0.99\linewidth]{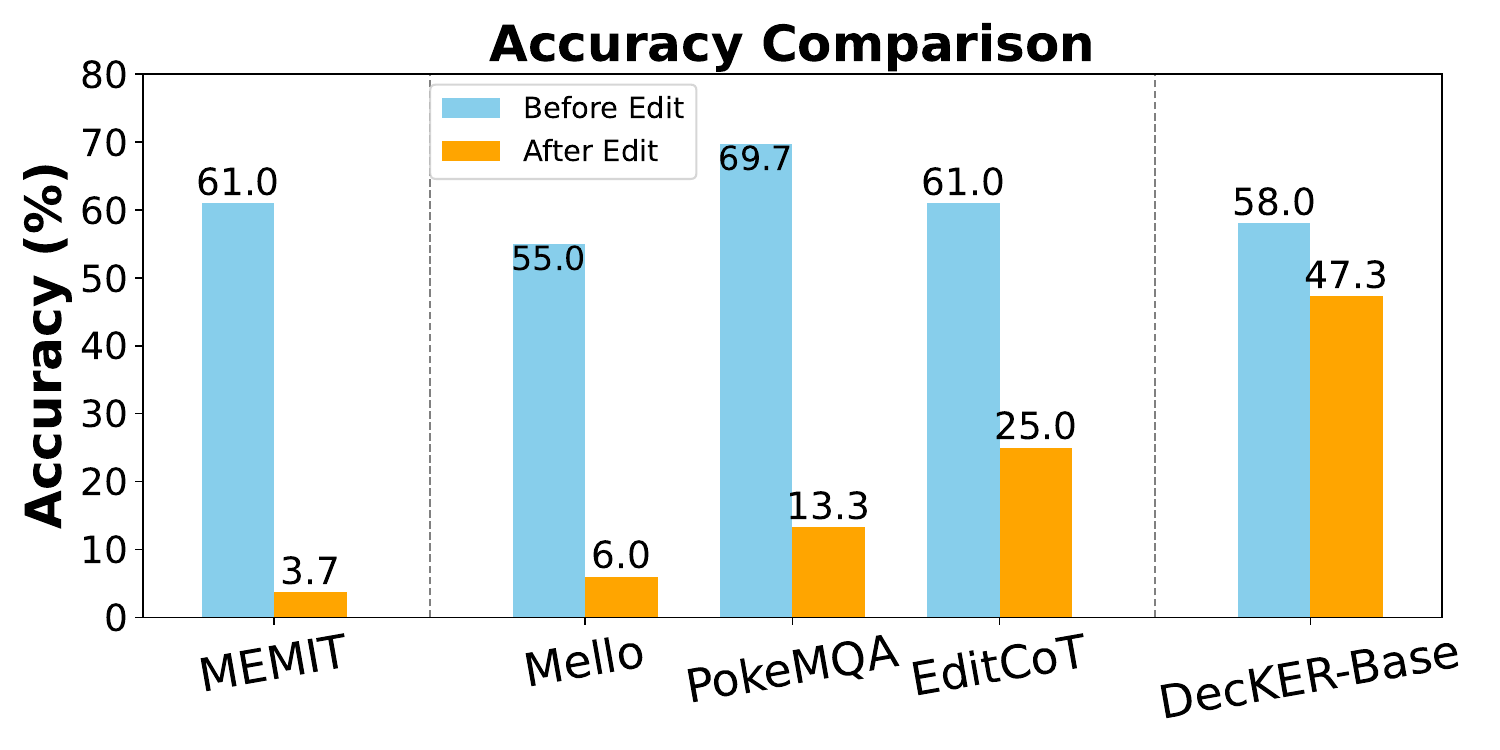}
\includegraphics[width=0.99\linewidth]{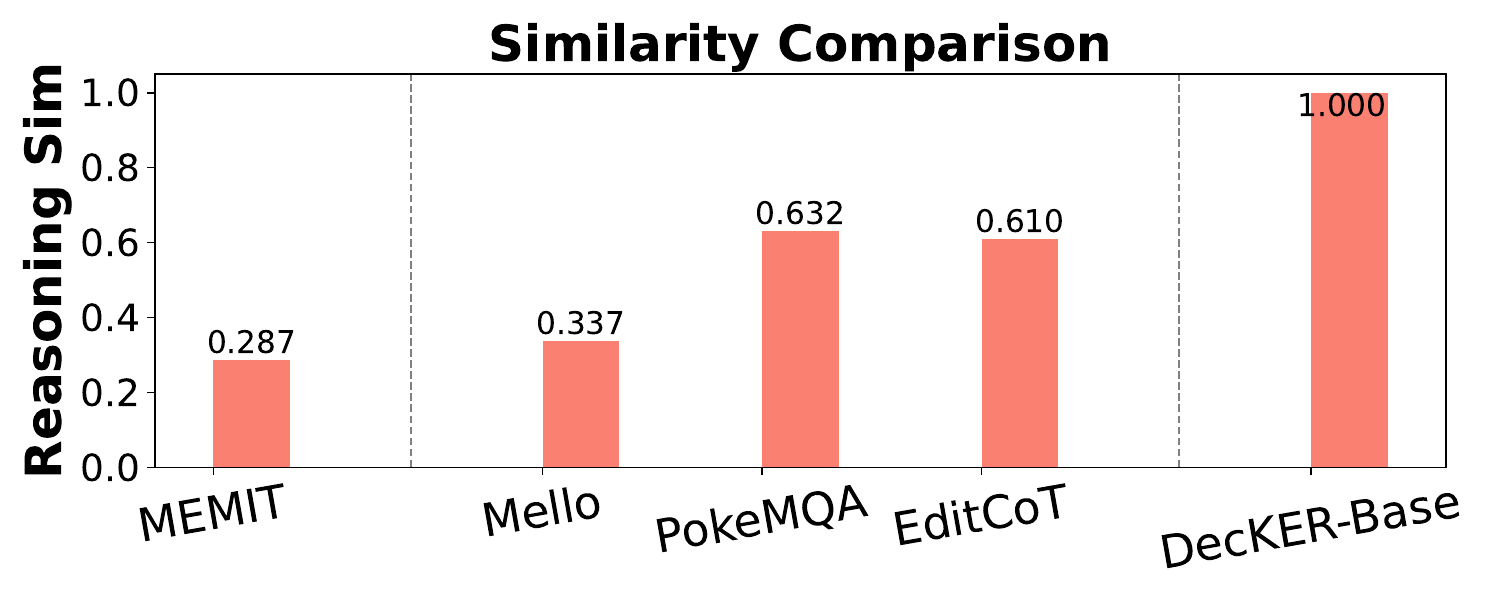}
        \caption{Performance and reasoning similarity of different KE Methods before and after editing. The second image indicates the similarity between the post-editing and pre-editing reasoning frameworks.}
        \label{fig:compare}
\end{figure}

\paragraph{Experimental Setup}

We randomly select 300 questions from MQuAKE-CF-3k-v2 \cite{zhong2024mquakeassessingknowledgeediting}, along with the associated edited facts, for our experiment. We apply a knowledge editing method to Llama-3.1-8B-Instruct \cite{dubey2024llama3herdmodels} and compare its performance before and after editing. We evaluate several methods, including MEMIT (a parameterized method), Mello, PokeMQA, and EditCoT. The first two are classic methods of parameterized knowledge editing and in-context editing, while the latter two represent the state-of-the-art ICE methods. The experiment consists of two main parts:

\textbf{Performance Comparison}: For multi-hop questions involving edited knowledge, the answers change before and after editing. We use the ground truths before and after editing as references, with accuracy as the metric, to evaluate performance before and after editing, respectively. For MEMIT, whether before or after editing, we prompt the LLM to generate a CoT during inference.
For EditCoT, we also get pre-editing answers through CoT reasoning.
For Mello and PokeMQA, pre-editing performance is obtained by replacing the edited fact in the retrieval corpus with the original one. 

\textbf{Reasoning Framework Comparison}: The reasoning framework $R_Q$ for a question is independent of the edit set $\mathcal{E}$, meaning the relation lists in the paths should remain unchanged after editing. A good ICE method should maintain this similarity to preserve the model’s reasoning ability. We use regular expressions to identify the reasoning paths generated by each knowledge editing method. For CoT-based methods, the CoT itself serves as the reasoning path, while for sub-question decomposition methods, we combine the answers to all sub-questions into the  reasoning path. We then use GPT-4o-mini-0718 \cite{openai2024gpt4technicalreport} to identify the relations in the knowledge triples involved at each reasoning step, forming the reasoning framework $R$. Prompts are detailed in \autoref{apd:prompt1}. 
    
Then, we compare the similarity of the reasoning framework before and after editing for each method. Given two reasoning frameworks $R_1 = [r_{11},\dots, r_{1n}]$ and $R_2 = [r_{21}, \dots, r_{2m}]$, we compute the similarity using the following formula:

    \vspace{-1.5mm}
    \begin{small}
        \begin{equation}
            Simlarity(R_1, R_2) = \frac{\sum_{i=1}^{min(n,m)} \text{Sim}(r_{1i}, r_{2i})}{max(n,m)} \ . \label{equ:sim}
        \end{equation}
        \end{small}
    
It calculates the similarity between corresponding elements of two ordered lists, padding the shorter list with zeros if they differ in length. In \autoref{equ:sim}, the "Sim" function employs jina-embeddings-v3\cite{sturua2024jinaembeddingsv3multilingualembeddingstask} to compute the embeddings of the two elements individually and calculates their cosine similarity.



\paragraph{Results and Analysis}

\autoref{fig:compare} presents the results. Both MEMIT and current ICE methods show significant performance degradation. 
Ideally, a knowledge editing method should maintain a reasoning framework similarity close to 1.00, but the tested methods do not exceed 0.65, demonstrating substantial reasoning degradation. 
This phenomenon is most severe for the parameterized method MEMIT, as noted in previous research \cite{li2024reallyeditlanguagemodels}. ICE, which does not modify LLM parameters, should theoretically maintain the model's reasoning capability, but the low reasoning framework similarity for the three ICE methods suggests that they fail to achieve this goal. Additionally, we observe a positive correlation between editing performance and reasoning framework similarity, further emphasizing the importance of maintaining the reasoning framework. In reasoning tasks, the reasoning planning and knowledge injection of current ICE methods are coupled, and the injection of conflicting knowledge affects reasoning planning. Therefore, this highlights the importance of decoupling reasoning from editing.

\section{Methodology}

\begin{figure*}[t]
        \centering
        \includegraphics[width=\linewidth]{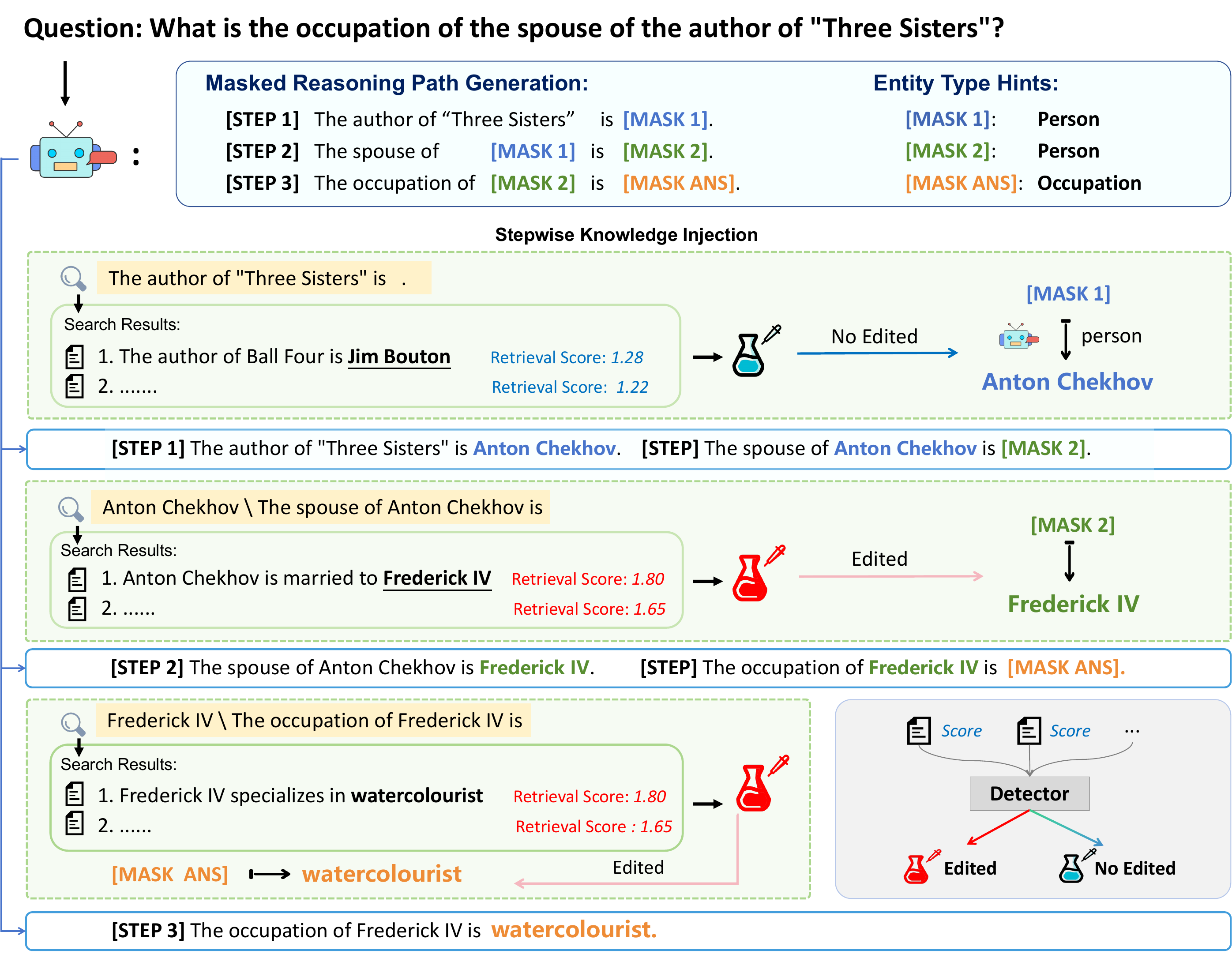}
        \caption{The workflow of DecKER, only showing the processing of a single path, omitting the final answer selection stage for multiple sampled reasoning paths.}
        \label{fig:dke}
\end{figure*}

We propose DecKER, a method that decouples reasoning and knowledge injection to maintain the reasoning framework while editing. The workflow is shown in \autoref{fig:dke}, with its pseudocode in \autoref{apd:pseudo}.


\subsection{Masked Reasoning Path Generation}

We separate reasoning from editing by initially having the LLM conduct global planning. Specifically, we first prompt the LLM to generate a masked reasoning path for the provided multi-hop question $Q$. 
"Masked" means that during generation, the LLM replaces all positions that require entity generation (except those already present in the question) with [MASK *] symbols, where * can be an integer or string.
We provide the LLM with a 5-shot prompt, leveraging the instruction-following and in-context learning capabilities of the LLM to accomplish this.
This approach aims to preserve high-quality reasoning paths from the original LLM before editing. In practice, we instruct the LLM to start each reasoning step with a [STEP] symbol.
Entities are numbered sequentially by the LLM to maintain consistency, as illustrated in \autoref{fig:dke}, where the author of "Three Sisters" is represented as [MASK 1] in the first two steps. The final answer is represented as [MASK ANS].

We also have the LLM provide entity type hints, such as "country", "company", etc. This is crucial because each filling step is completed independently. Without type hints, the model might generate completions that fit the instructions but do not align with the reasoning framework. All of the prompt templates are detailed in \autoref{apd:prompt1}.

\subsection{Stepwise Knowledge Injection}

Then we fill in each masked entity step by step.

\noindent \textbf{1. Retrieval}: For a masked step currently being filled, we remove the [MASK *] tag in it and use it as a query to search within the edited memory (i.e., the union set of all $\mathcal{E}$ in the dataset). If previous entities have been filled, we include the most recent entity in the query.

\noindent \textbf{2. Conflict Detection}: Conflict detection is performed by comparing the retrieval results with the query, using a hybrid method that combines retrieval scores and LLM judgments. When a query involves edits, the retrieval scores exhibit two key features: a notably high score for the top result and a substantial gap between the highest and second-highest scores, mathematically represented as:

\vspace{-1.5mm}
\begin{small}
\begin{subequations}
\begin{equation}
S(d_1) > \alpha \ , \label{equ:alpha}
\end{equation}
\begin{equation}
S(d_1) - S(d_2) > \beta \ , \label{equ:beta}
\end{equation}
\end{subequations}
\end{small}

\noindent where $S$ is the retrieval score. In our experiments, it denotes the dot-product similarity between the embeddings of documents and queries, with \(d_i\) representing the result ranked \(i\). \(\alpha\) and \(\beta\) are hyperparameters. If both criteria yield consistent results, we accept them without further LLM evaluation. Otherwise, LLM analysis is conducted.
Each edited fact maps to an original one, allowing the LLM to determine if the original supports the masked step's filling. Support indicates a conflict between the edited fact and the masked step.

\noindent \textbf{3. Filling}: When a retrieval result conflicts with the masked step, we directly replace the masked part with the object from the knowledge triple corresponding to the retrieval result, and also update subsequent identical masked parts. For example, in \autoref{fig:dke}, all instances of [MASK 2] are replaced with "Frederick IV". If no conflict is detected, the model treats the masked part as a fill-in-the-blank task, guided by the previously assigned entity type. This process continues until [MASK ANS] is filled, providing the final answer to the question.

\subsection{Final Answer Selection}

In the workflow, we assume that each masked step involves only one entity to be filled, which is reasonable for the MQA task. However, due to insufficient few-shot learning ability, the model might generate incorrect or incomplete masked reasoning paths, and the filled entities might not align with the intended types.
To ensure the quality of the final answers, we filter multiple candidate answers generated through the above process with two evaluation methods — reasoning path planning and knowledge injection — to select the final answer.

 \noindent \textbf{1. Reasoning Path Planning Evaluation (RPP Eval)}: Inspired by the work on hallucinations\cite{su-etal-2024-unsupervised,su2024mitigating} in LLMs, we can evaluate the reasoning path by calculating the model's uncertainty, as LLMs are prone to hallucinations or incorrect outputs when uncertain. For each generated token, Predictive Entropy (PE) is defined as follows:

\vspace{-1mm}
\begin{small}
\begin{equation}
PE = -\sum_{\tilde{w} \in W} p_i(\tilde{w}) log \ p_i(\tilde{w}) \ , \label{equ:rpp}
\end{equation}
\end{small}

\noindent where $W$ is the vocabulary of the LLM, and $p_i(\tilde{w})$ is the generation probability of token $\tilde{w}$ at this step. For each masked path, we use the negative average PE over all generated tokens as the evaluation score, with higher scores indicating greater certainty.

 \noindent \textbf{2. Knowledge Injection Evaluation (KI Eval)}: After filling the reasoning path, we obtain a list of filled entities. For each filled entity $e_i$, we instruct the model to determine whether it matches the pre-assigned type $t_i$. The score is calculated as follows:

\vspace{-1mm}
\begingroup
\fontsize{8.5}{5}\selectfont
\begin{equation}
KI(e_i, t_i) = \mathds{1} \bigl[ P_M(\text{``yes''} \mid e_i, t_i) > P_M(\text{``no''} \mid e_i, t_i) \bigr] \ , \label{equ:ki}
\end{equation}
\endgroup

\noindent where $M$ is the LLM and $\mathds{1}$ is an indicator function. The prompt template is detailed in \autoref{apd:prompt1}. We compute the KI Eval score by averaging the scores of all filled entities. In practice, RPP Eval is used to retain the top 50\% of sampled masked reasoning paths for filling. The final answer is selected from the filled path with the highest KI Eval score.

\subsection{Discussion}

We conduct the experiments in \autoref{sec:sec2.2} on DecKER. For the pre-editing settings, we replace the edited memory with the corresponding original facts. The results indicate DecKER exhibits only a slight accuracy drop post-editing and significantly outperforms other baselines. Additionally, DecKER maintains complete consistency in reasoning frameworks between pre- and post-editing, as they share the same masked reasoning path, highlighting the reason behind its strong performance.

\begin{table*}[t]
\centering
\caption{The overall results. Ripple-Pop and Ripple-Rand denote the popular and random subsets of RippleEdits. The metric is Multi-Hop Accuracy (\%). We bold the top performing methods and underline the second-best ones.}
\resizebox{0.85\textwidth}{!}{
\begin{tabular}{llcccc}
\toprule
\textbf{Models} &
  \textbf{Methods} &
  \multicolumn{1}{l}{\textbf{MQuAKE-CF-3k-v2}} &
  \multicolumn{1}{l}{\textbf{MQuAKE-T}} &
  \multicolumn{1}{l}{\textbf{Ripple-Pop}} &
  \multicolumn{1}{l}{\textbf{Ripple-Rand}} \\ \midrule
\multirow{9}{*}{\textbf{Llama-3.1-8B-Instruct}} & \textbf{MEMIT}       & 6.8           & 44.1          & 8.4           & 9.1           \\
                                                & \textbf{AlphaEdit}   & 7.0           & 42.0          & 8.6           & 9.6           \\
                                                & \textbf{Mello}       & 11.0          & 57.7          & 25.8          & 32.7          \\
                                                & \textbf{PokeMQA}     & 16.8          &  77.8& 30.4& 33.1\\
                                                & \textbf{DeepEdit}    & 8.9& 57.5& 3.4& 5.0\\
                                                & \textbf{RAE}         &  54.3& 50.9          & 20.6          & 17.0          \\
                                                & \textbf{EditCoT}     & 36.0& 74.6&  31.6& 22.3          \\ \cmidrule{2-6} 
                                                & \textbf{DecKER-Base}     & {\ul 57.8}& {\ul 80.9}& {\ul 46.6}& \textbf{46.1}\\
                                                & \textbf{DecKER-BoN} \ ($N=6$) & \textbf{59.0}& \textbf{81.3}& \textbf{47.4}& {\ul 44.9}\\ \midrule
\multirow{9}{*}{\textbf{Qwen2.5-7B-Instruct}}   & \textbf{MEMIT}       & 6.6           & 21.5          & 9.8           & 17.2          \\
                                                & \textbf{AlphaEdit}   & 7.1           & 21.1          & 11.0          & 17.0          \\
                                                & \textbf{Mello}       & 1.7           & 50.3          & 3.6           & 1.4           \\
                                                & \textbf{PokeMQA}     & 7.1           & 45.3          & 7.0           & 7.6           \\
                                                & \textbf{DeepEdit}    & 12.3          & 25.3          & 6.0           & 12.0          \\
                                                & \textbf{RAE}         &  38.3& 28.7          & 24.0          & 18.3          \\
                                                & \textbf{EditCoT}     & 26.2&  \textbf{74.5} &  29.2& 30.7\\ \cmidrule{2-6} 
                                                & \textbf{DecKER-Base}     & {\ul 42.1}&  67.7& {\ul 35.8}& {\ul 32.1}\\
                                                & \textbf{DecKER-BoN} \ ($N=6$) & \textbf{49.8}& {\ul 70.8}& \textbf{39.8} & \textbf{38.3}\\ \midrule
\multirow{6}{*}{\textbf{Qwen2.5-14B-Instruct}}  & \textbf{Mello}       & 1.2           & 26.4          & 4.6           & 2.3           \\
                                                & \textbf{PokeMQA}     & 2.3           & 32.4          & 5.9           & 3.9           \\
                                                & \textbf{DeepEdit}    & 7.3           & 51.7          & 9.4           & 14.4          \\
                                                & \textbf{EditCoT}     & 37.0& {\ul 82.7}& 35.0&  33.5\\ \cmidrule{2-6} 
                                                & \textbf{DecKER-Base}     & {\ul 50.8}&  81.4& \textbf{45.0}& {\ul 45.6}\\
                                                & \textbf{DecKER-BoN} \ ($N=6$) & \textbf{56.6}& \textbf{83.6}& {\ul 44.0}& \textbf{46.1}\\ \bottomrule
\end{tabular}
}
\label{tab:maintab}
\end{table*}

\section{Experimental Settings}

\subsection{Evaluation Details}
We evaluate our methods using four datasets: MQuAKE-CF-3k-v2, MQuAKE-T\cite{zhong2024mquakeassessingknowledgeediting}, and the Popular and Random subsets of RippleEdits\cite{cohen2023evaluatingrippleeffectsknowledge}. MQuAKE-CF-3k-v2 comprises 3,000 questions with 2,764 edits, with 2-, 3-, and 4-hop questions. MQuAKE-T contains 1,868 questions and 96 edits, with 2- and 3-hop questions. RippleEdits-Popular has 500 2-hop questions and 266 edits, while RippleEdits-Random has 1,137 2-hop questions and 626 edits.

As in the original paper of Mello, we employ the Multi-Hop Accuracy metric, where a question is correct if at least one rewritten version is answered correctly. For MQuAKE datasets, each test includes 3 rewritten questions, while RippleEdits datasets contain only 1 question per test.

Three LLMs are evaluated: Meta-Llama-3.1-8B-Instruct\cite{dubey2024llama3herdmodels}, Qwen2.5-7B-Instruct, and Qwen2.5-14B-Instruct\cite{qwen2.5}, using their official chat templates across all baselines. 

\subsection{Baselines}
We compare DecKER with other ICE methods and two parameterized approaches: MEMIT\cite{meng2023masseditingmemorytransformer} and AlphaEdit\cite{fang2024alphaeditnullspaceconstrainedknowledge}. These two approaches focus on editing neurons related to the targeted knowledge, with AlphaEdit mitigating unintended disruptions through matrix projection. ICE methods include: (1)  Mello\cite{zhong2024mquakeassessingknowledgeediting} and PokeMQA\cite{gu2024pokemqaprogrammableknowledgeediting}, subproblem-based methods;  (2) DeepEdit\cite{wang2024deepedit}, depth-first search-based methods; and (3) EditCoT\cite{wang2024knowledgeeditingchainofthought}, CoT editing methods. We also evaluate RAE\cite{shi24rae}, a method that achieves knowledge editing by modifying knowledge graphs.
Due to the resource-intensive nature of the parameterized methods and RAE, we test them only on the two smaller LLMs. Detailed implementation is in \autoref{apd:baselines}.

\subsection{Implementation Details of DecKER}

We evaluate two settings:

\textbf{DecKER-BoN}: Sample $N$ masked reasoning paths, use RPP Eval (\autoref{equ:rpp}) to select the best half for filling, and then employ KI Eval (\autoref{equ:ki}) to choose the best result. One greedy decoding is followed by $N-1$ top-p sampling (probability threshold 0.95, temperature 1.2). Other generation processes use greedy decoding. In our experiments, $N$ is set to 6.

\textbf{DecKER-Base}: This method does not utilize two-round evaluation. Instead, it generates one masked reasoning path using greedy decoding. If the format is incorrect, e.g., no masked parts, a new path is sampled by nucleus sampling. Failure in both attempts is viewed as an error result.

In conflict detection, $\alpha$ is set to 1.5 and $\beta$ is 0.1.

\subsection{Retrieval Settings}
\label{sec:settings}
Except for RAE, which retrieves from a knowledge graph, all ICE methods adopt the same retrieval approach following Mello, using contriever-msmarco~\cite{izacard2022unsuperviseddenseinformationretrieval} as the retriever to access a knowledge base of edited knowledge.
Notably, RAE accesses Wikidata\cite{wang2021kepler} during inference steps, even involving unedited knowledge, creating an unfair comparison. 
To ensure consistency, we modify its graph retrieval process: if the subject and the retrieved relation involve unedited knowledge triples, the LLM generates the object of the triple instead of getting it from Wikidata. This adjustment prevents data leakage, ensuring a uniform retrieval scope across all methods.

\begin{table}[t]
\centering
\caption{Results on GPT-4o-mini. The dataset is MQuAKE-300. The best is in bold.}
\resizebox{0.45\textwidth}{!}{
\begin{tabular}{ccccc}
\toprule
\textbf{Mello} & \textbf{PokeMQA} & \textbf{EditCoT} & \textbf{RAE} & \textbf{DecKER-Base} \\ \midrule
11.0           & 51.3             & 46.7         & 67.0             & \textbf{69.7}             \\ \bottomrule
\end{tabular}
 }
\label{tab:gpt}
\end{table}
\vspace{-1mm}

\begin{figure*}
  \begin{minipage}[t]{0.34\textwidth}
    \includegraphics[width=\linewidth]{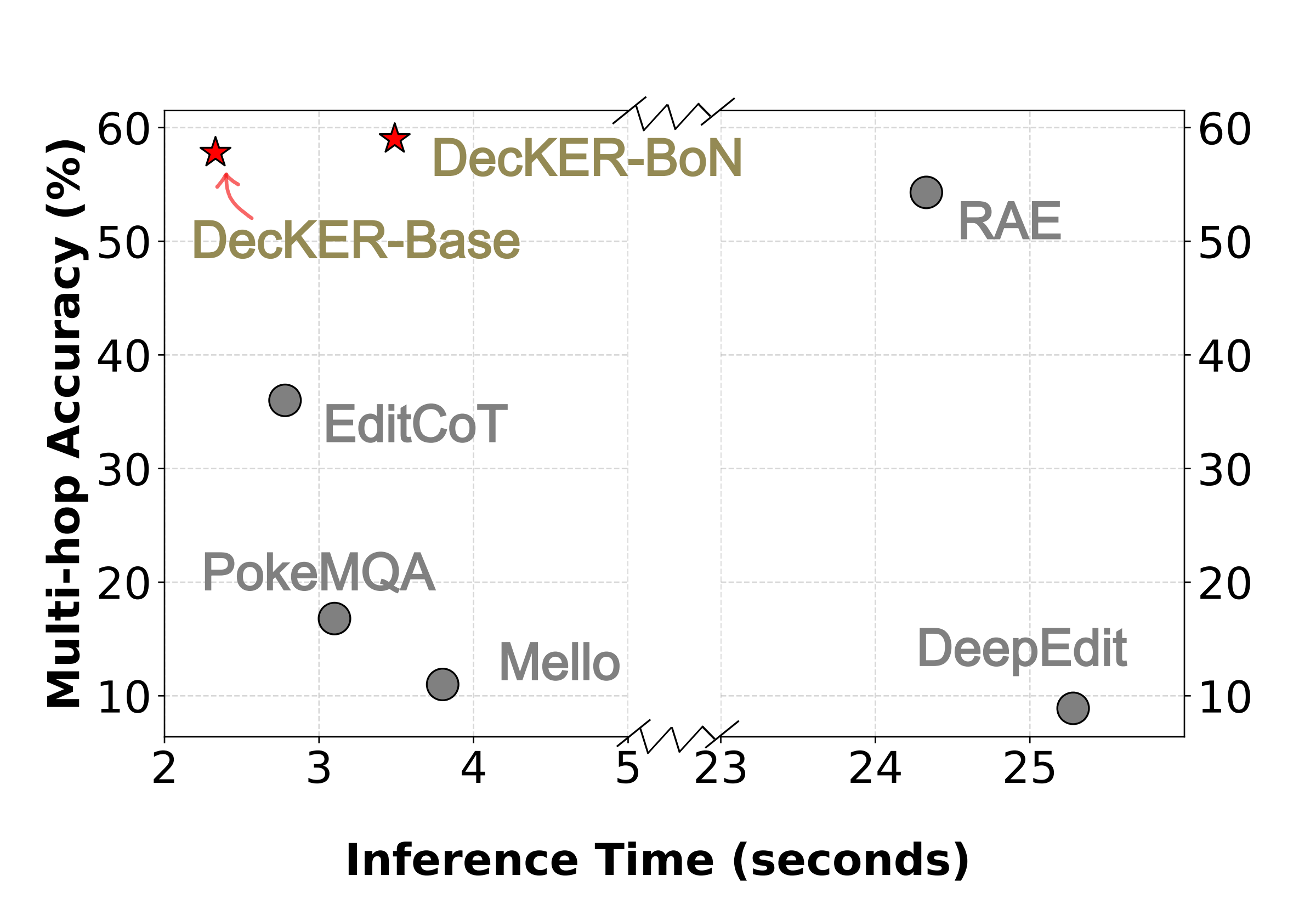}
    \caption{Avg. Time vs. Performance. We truncate the x-axis due to its length.}
    \label{fig:efficiency}
  \end{minipage}
  \hfill
  \begin{minipage}[t]{0.63\textwidth}
    \begin{minipage}[t]{0.49\linewidth}
      \includegraphics[width=\linewidth]{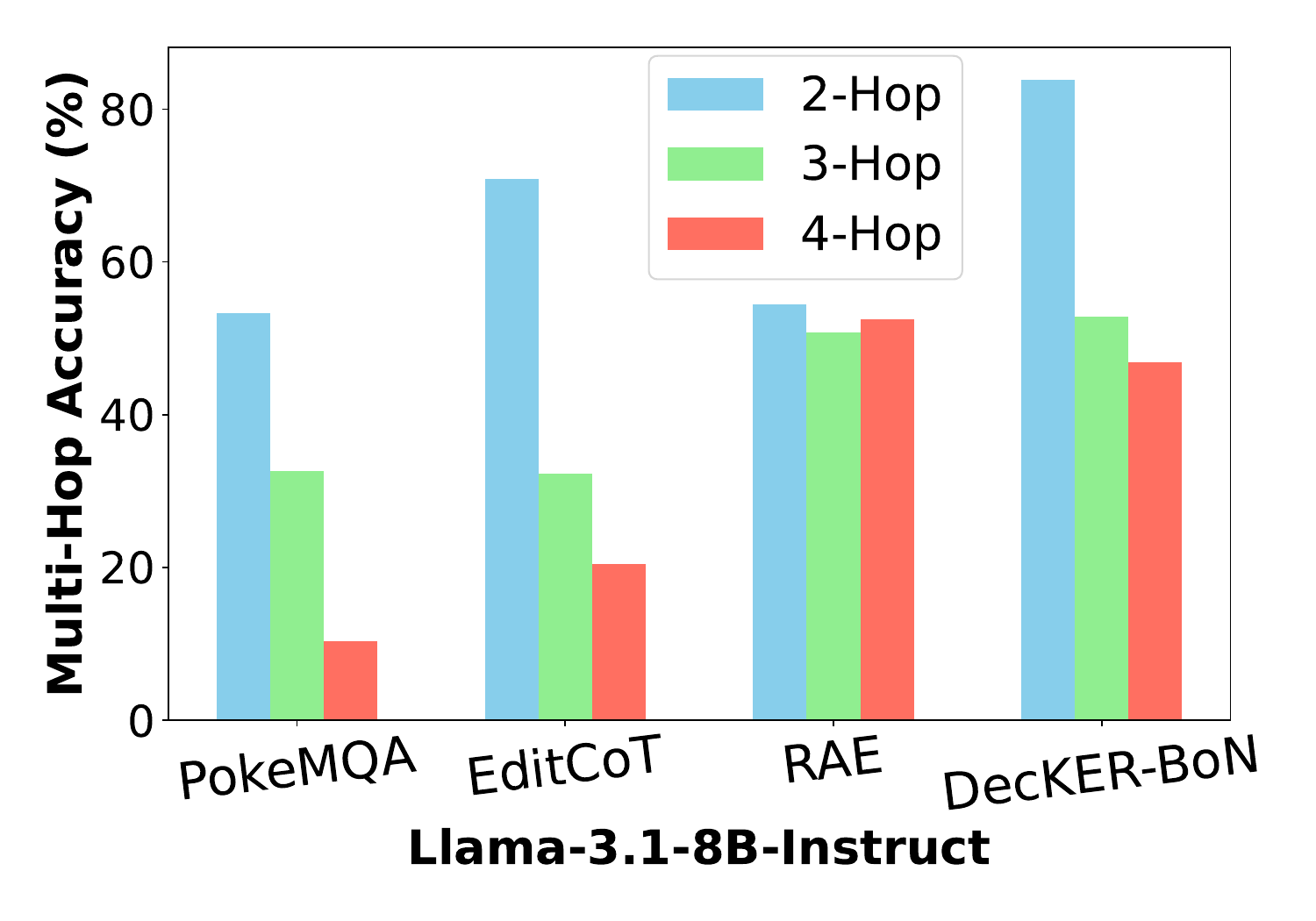}
    \end{minipage}
    \hfill
    \begin{minipage}[t]{0.49\linewidth}
      \includegraphics[width=\linewidth]{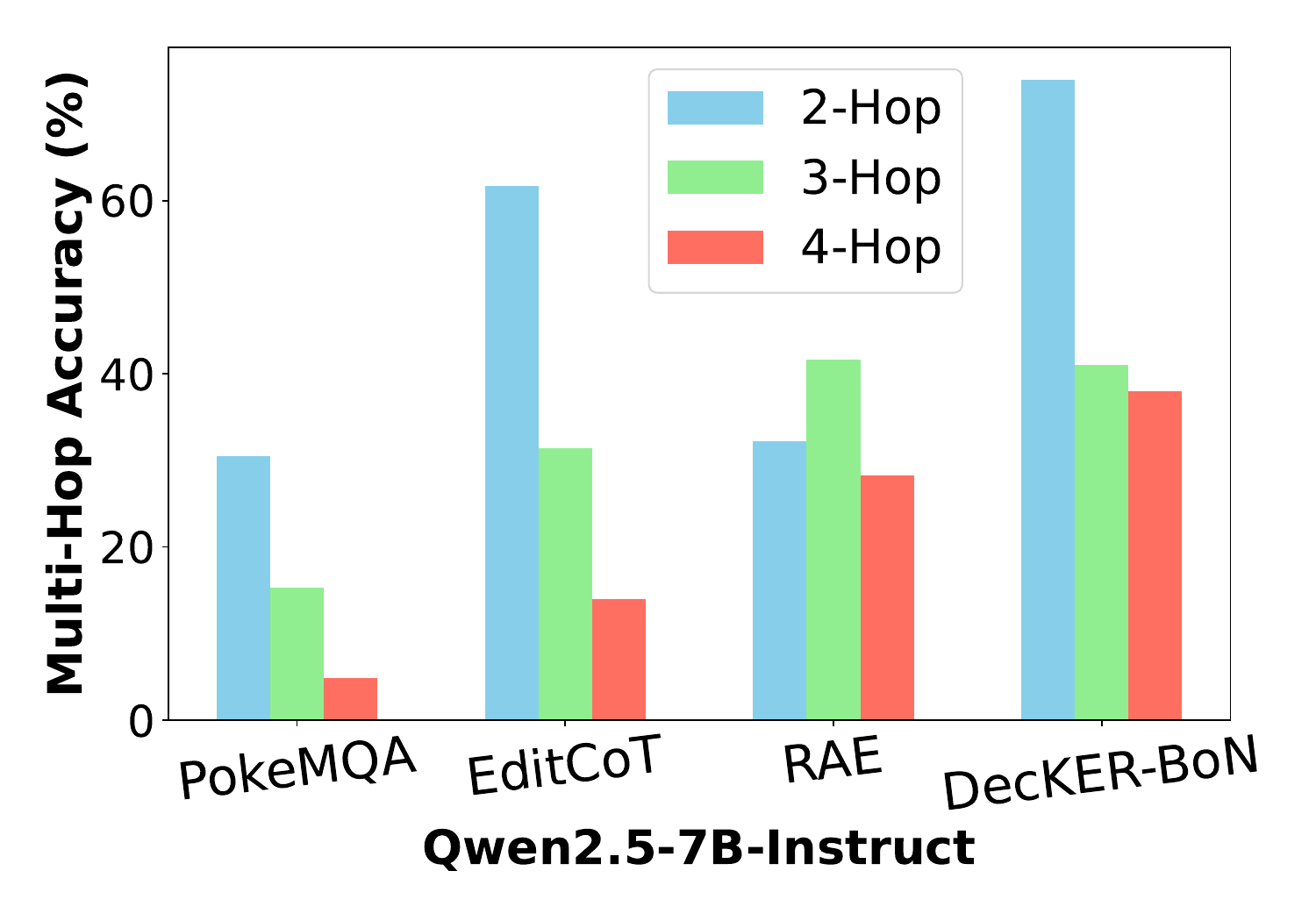}
    \end{minipage}
    \caption{\#Hop vs. Performance: performance on the MQuAKE-CF-3k-v2 and -T datasets based on the number of hops of each problem.}
    \label{fig:hop_num}
  \end{minipage}
\end{figure*}

\section{Experimental Results}

\subsection{Main Results}
\textbf{DecKER outperforms all baselines.} \autoref{tab:maintab} presents the main results, showing that DecKER-Base and DecKER-BoN outperform previous methods across most models and datasets, particularly excelling in the Popular subset of RippleEdits. This highlights DecKER's effectiveness in handling multi-hop reasoning and the Ripple Effect after editing. Moreover, DecKER demonstrates consistent performance, unlike other baselines, such as PokeMQA, which may excel in some datasets and models but perform poorly in others. This suggests that DecKER achieves a stable decoupling of reasoning and editing, maintaining the model's original reasoning capabilities.

\textbf{Sampling multiple paths enhances DecKER's performance.} DecKER-BoN outperforms the Base version by selecting reasoning paths through two evaluation rounds from options, prioritizing paths with higher LLM confidence and better entity type matching. However, the BoN version occasionally underperforms the Base version, suggesting the potential of improving evaluation design.


\textbf{Preserving the reasoning framework is crucial.} EditCoT and PokeMQA exhibit more competitive performance than Mello and MEMIT. As analyzed in \autoref{sec:sec2.2}, despite these two methods not being able to completely retain the pre-edit reasoning framework, they still surpass Mello and MEMIT in the similarity dimension, leading to higher answer accuracy. RAE, a graph retrieval method, constrains the search space for the construction of reasoning frameworks by LLMs, as the edges in the graph are fixed, resulting in commendable performance. These results confirm that maintaining the reasoning framework is essential.

\subsection{Comparative Analysis}
\textbf{Performance on Proprietary LLMs} One advantage of in-context editing is its applicability to proprietary models. We test GPT-4o-mini-0718. Following \citet{shi24rae}, we randomly select 300 questions from MQuAKE-CF-3k-v2 along with the corresponding edits, named MQuAKE-300. Since some processes in EditCoT and RAE, such as the editor or the complete output probabilities, cannot be executed on proprietary LLMs, we use Llama-3.1-8B-Instruct as a proxy model, following  \citet{wang2024knowledgeeditingchainofthought}. The results are presented in \autoref{tab:gpt}. DecKER-Base outperforms all baselines, indicating its potential for editing proprietary LLMs.

\textbf{Efficiency} \autoref{fig:efficiency} presents the average inference time per problem and performance on MQuAKE-3k-CF-v2 for various ICE methods. The LLM is Llama-3.1-8B-Instruct. DecKER-Base is positioned in the upper left corner, achieving superior editing results over other baselines in the shortest time. DecKER-BoN, due to the additional evaluation and more sampled sentences, takes slightly longer but achieves the best. Among all the baselines, the top-performing RAE requires over 24 seconds per problem, while the fastest EditCoT exhibits weaker performance compared to ours, highlighting the efficiency advantage of our method. The details are in \autoref{apd:efficiency}.

\textbf{Performance on Problems with Different Hop Numbers} We analyze the performance of several competitive methods on two MQuAKE datasets based on the number of hops, as shown in \autoref{fig:hop_num}. DecKER-BoN consistently ranks first or second in 2-4 hop problems, with particularly strong performance on 2-hop problems, reflecting the inherent reasoning abilities of LLMs. RAE leverages a knowledge graph, maintaining stable performance across 2-4 hops by constraining the exploration space during inference with fixed graph edges. Despite this, DecKER matches or exceeds RAE's performance on 3 and 4 hop problems.

\subsection{Ablation Studies}

\begin{table}[t]
\centering
\caption{An ablation study on the selection process on MQuAKE-CF-3k-v2. Using "L" for Llama-3.1-Instruct and "Q" for Qwen2.5-Instruct. "Random Choice" refers to randomly selecting one answer from 6 sampled paths.}
\resizebox{0.85\linewidth}{!}{
\begin{tabular}{llll}
\toprule
                          & \textbf{L-8B} & \textbf{Q-7B} & \textbf{Q-14B} \\ \midrule
\textbf{DecKER-BoN}  & 59.9          & \textbf{49.8}& \textbf{56.6}\\
\textit{\qquad w/o RPP Eval}     & \textbf{60.2}& 49.4          & 56.2           \\
\textit{\qquad w/o KI Eval}      & 58.1          & 49.5          & 55             \\ \midrule
\textbf{DecKER-Base} & 57.8          & 42.1          & 50.8           \\ \midrule
\textbf{Random Choice}    & 55.2          & 47.1          & 51.6           \\ \bottomrule
\end{tabular}
}
\label{tab:ab1}
\end{table}

In this section, We perform ablation studies on Llama-3.1-8B-Instruct.

\begin{figure}[!t]
        \centering
        \includegraphics[width=\linewidth]{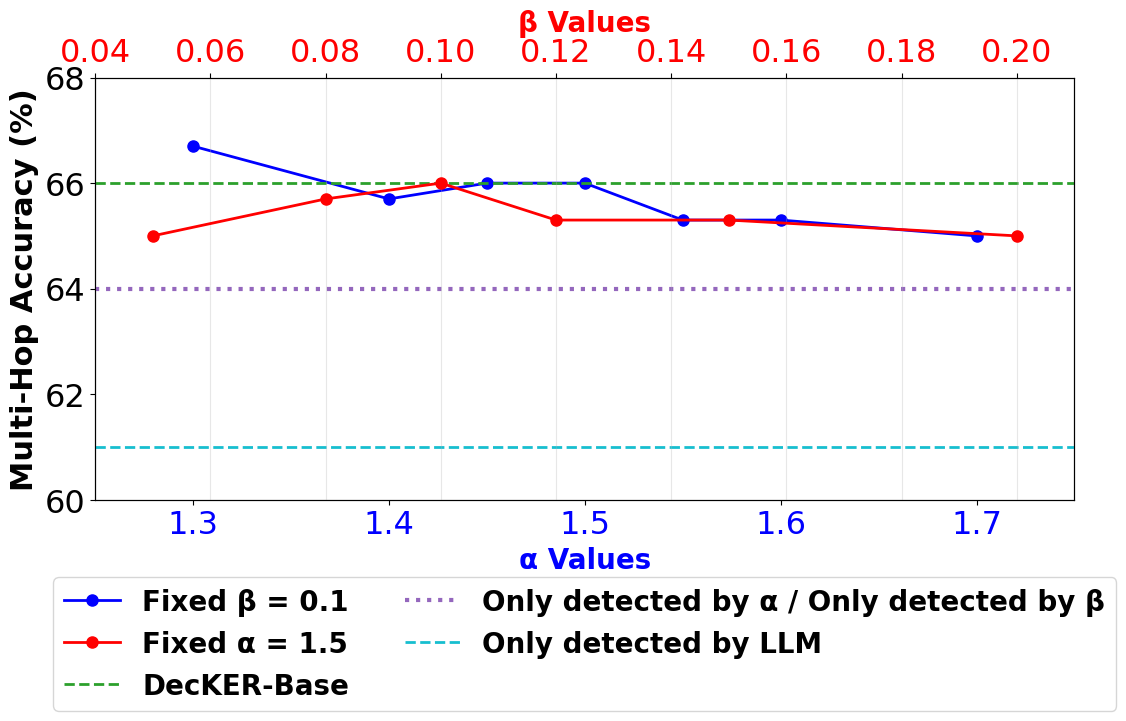}
        \caption{Ablation studies for conflict detection on DecKER-Base. The dashed lines compare single versus composite detection, with identical hyperparameters to the main experiments. The solid line shows the effect of varying one hyperparameter while fixing the other, with the color of x-axis labels matching the solid line.}
        \label{fig:ab_conflict}
\end{figure}

\textbf{Components of Conflict Detection} In \autoref{fig:ab_conflict}, we compare the performance of DecKER-Base with three experimental settings on MQuAKE-300: conflict detection 1) using only LLM; 2) using only the highest retrieval score as shown in \autoref{equ:alpha} (\textit{Only detected by $\alpha$}); 3) using only the difference between retrieval scores of the top two documents as shown in \autoref{equ:beta} (\textit{Only detected by $\beta$}). The hyperparameter settings are consistent with the main experiments. Results indicate that single-method detection degrades performance, especially with LLM-only detection. The combination of all three methods achieves the best performance. Additionally, we analyze hyperparameter impact by fixing either $\alpha$ or $\beta$ while varying the other, as shown in \autoref{fig:ab_conflict}. The stability of performance within a certain range of these parameters demonstrates the robustness of conflict detection.

\textbf{Two-round Evaluation} In \autoref{tab:ab1}, we present the impact of RPP eval and KI eval in DecKER-BoN on MQuAKE-CF-3k-v2. The Reasoning Path Planning Evaluation alone yields the poorest performance, while the Knowledge Injection Evaluation alone performs comparably to using both evaluations. Notably, RPP Eval involves filtering multiple masked reasoning paths, thus, the computational load during filling is less than that of performing only KI Eval, demonstrating its effectiveness in reducing computational costs. Additionally, selecting one answer randomly from 6 sampled paths significantly underperforms compared to DecKER-BoN. Additionally, we explore the impact of edit batch size and the number of sampled masked paths on our method. The results are shown in \autoref{apd:abs}.

\section{Conclusion and Future Works}
In this paper, we explore Multi-hop QA to highlight the importance of decoupling reasoning and knowledge injection when applying in-context editing. Consequently, we propose a novel in-context editing method, DecKER, which achieves this decoupling by first planning the reasoning path and then filling knowledge entities. Our method demonstrates strong performance across 3 LLMs and 4 datasets, offering a new perspective for optimizing in-context editing methods.

In future work, we plan to extend DecKER to a broader range of tasks beyond multi-hop QA, such as fact verification~\cite{thorne2018fever,fvsurvey} and long-form text generation~\cite{su2025judge,bai2023longbench}, where reasoning quality and factual accuracy are both critical. 
Another promising direction is examining the connection between our findings and Retrieval-Augmented Generation (RAG)~\cite{lewis2020retrieval,dong2025decoupling,su2024dragin,su2025parametric,tu2025rbft,su2024wikiformer}. Similar to ICE, RAG injects retrieved content into the model input, risking semantic conflicts with parametric knowledge. Inspired by our preliminary analysis, we aim to adapt DecKER to the RAG setting, investigating whether decoupling retrieval-based knowledge injection from reasoning path planning can enhance generation consistency and robustness.

\section{Limitations}
In this work, we conduct evaluations using three widely-used open-source LLMs with parameter sizes ranging from 7B to 14B, as well as a proprietary LLM. Due to resource constraints, we do not perform experiments on larger parameter-scale open-source models. Additionally, experiments with the three resource-intensive baselines are conducted only on the 7B and 8B parameter LLMs.

Besides, we focus on the enhancement of the knowledge editing process, with DecKER's performance relying solely on the LLM's own reasoning capabilities. The integration of knowledge augmentation techniques, such as knowledge graphs, to further improve DecKER's performance remains an open area of research.

\bibliography{custom}

\appendix

\section{Pseudocode Description of DecKER-BoN}
\label{apd:pseudo}
The pseudocode of DecKER-BoN is shown in Algorithm 
 \autoref{alg:decouple_ke}.

\begin{algorithm*}
\caption{DecKER}
\label{alg:decouple_ke}
\begin{algorithmic}[1]
\Require Question $Q$, LLM $M$, Knowledge Edits $\mathcal{E}$, Hyperparameters $\alpha$, $\beta$, Sample Size $N$
\Ensure Final Answer $A$

\State \textbf{Step 1: Generate Masked Reasoning Paths and Entity Types}
\For{$i \gets 1$ to $N$}
    \State $R_Q^{(i)}, \{t_1, t_2, \dots \} \gets \textsc{GenerateMaskedReasoningPath}(M, Q)$
\EndFor

\State \textbf{Step 2: Fill Masked Entities}
\For{each masked reasoning path $R_Q^{(i)}$}
    \State PathLength $\gets \text{length}(R_Q^{(i)})$
    \State $P_Q^{(i)} \gets [R_Q^{(i)}[0]]$ 
    \For{$j$ in $[0, 1, \dots, PathLength-1]$}
        \State $s_k \gets P_Q^{(i)}[j]$ \Comment{The current masked step}
        \State MaskTag $\gets \text{GetMaskTag}(s_k)$
        \State Query $\gets \text{PrevFilledEntity}() +  \text{RemoveMaskTag}(s_k)$
        \State $\{(d_1, score_1), (d_2, score_2), ...\} \gets \textsc{Retrieve}(\mathcal{E}, \text{Query})$
        \State IsEdited $\gets False$
        \If{$(score_1 > \alpha) = (score_1 - score_2 > \beta)$}
            \State IsEdited $\gets (score_1 > \alpha)$
        \Else
            \State IsEdited $\gets \textsc{ConflictCheck}(M, s_k, \text{OriginalFact}(d_1))$
        \EndIf

        \If{IsEdited}
        \State $e_k \gets \text{GetObjectEntity}(d_1)$
        \Else
        \State $e_k \gets M(\text{"Fill \{MaskTag\} with type } t_k\text{"})$ \Comment{LLM-based completion}
        \EndIf
        \State Replace all MaskTag in $R_Q^{(i)}$ with $e_k$
        
        \If{MaskTag = "[MASK ANS]"}
        \State $A^{(i)} \gets e_k$
        \State break
        \EndIf
        \State $P_Q^{(i)} \gets P_Q^{(i)} + [R_Q^{(i)}[j+1]]$
    \EndFor
    
\EndFor
\State \textbf{Step 3: Select the Best Sample}
\State $\text{RPP Scores} \gets - \ 
\textsc{ComputeAvgPE}(M, \{R_Q^{(i)}\})$ 
\State $\text{KI Scores} \gets [ \ ]$
\State $\text{Answers} \gets [ \ ]$
\State $\text{TopPaths} \gets \text{Top 50\% } P_Q \text{ by RPP Scores}$
\For{each filled reasoning path $P_Q^{(i)}$ in TopPaths}
\State $\text{KI Scores}.append(\frac{1}{\text{length}(P_Q^{(i)})}\sum \mathds{1}[P_M(\text{"yes"}|\text{"Is } e_j \text{ type } t_j?\text{"}) > P_M(\text{"no"}|\text{"Is } e_j \text{ type } t_j?\text{"}) ])$
\State $\text{Answers}.append(A^{(i)})$
\EndFor
\State $A \gets \text{Answers}[\arg\max \text{KI Scores}]$

\end{algorithmic}
\end{algorithm*}

\section{Details of Baselines}
\label{apd:baselines}
\textbf{MEMIT \cite{meng2023masseditingmemorytransformer} and AlphaEdit \cite{fang2024alphaeditnullspaceconstrainedknowledge}:} We adopt the official implementation of AlphaEdit, setting the target editing layers to 4-8, with a learning rate of 0.1 and a weight decay of 0.5. The nullspace threshold for AlphaEdit is set to 0.02.

\textbf{Mello \cite{zhong2024mquakeassessingknowledgeediting}:} We adhere to the original paper's settings, decomposing up to 4 subproblems.

\textbf{PokeMQA \cite{gu2024pokemqaprogrammableknowledgeediting}:} We follow the original paper's settings, decomposing up to 5 subproblems. As the original repository does not provide entities for the new dataset, we allow the model to extract problem entities in a similar format.

\textbf{DeepEdit \cite{wang2024deepedit}:} In line with the official implementation, we set the search width limit to 2 and the depth limit to 9.

\textbf{RAE \cite{shi24rae}:} We use the code provided by the official implementation and adjust its retrieval process as described in \autoref{sec:settings} to ensure a fair comparison.

\textbf{EditCoT \cite{wang2024knowledgeeditingchainofthought}: } We adhere to the original paper's settings, conducting up to 4 rounds of CoT editing. The retrieval process is configured to operate solely on edit memory, as one of the official implementations, to ensure a fair comparison.

For all ICE methods, we use the official prompt templates and adapt them to fit the chat template format of each LLM. The LLMs we used are implemented by Huggingface Transformers library \cite{wolf-etal-2020-transformers}.

\section{Details of Efficiency Experiments}
\label{apd:efficiency}
All efficiency experiments are conducted on a single NVIDIA A100 (40G) GPU. We exclude network latency from the reported inference time of RAE. In accordance with the official implementation, RAE employs float32 floating point numbers, while other methods use float16, as we observe a considerable performance degradation in RAE with half-precision. In our implementation, many operations in DecKER use batch processing, including the sampling of multiple masked reasoning paths and parallel filling processes.

\section{Further Ablation Studies}
\label{apd:abs}
\subsection{Performance with Different Edit Batch Sizes}

\begin{figure}[t]
        \centering
        \includegraphics[width=\linewidth]{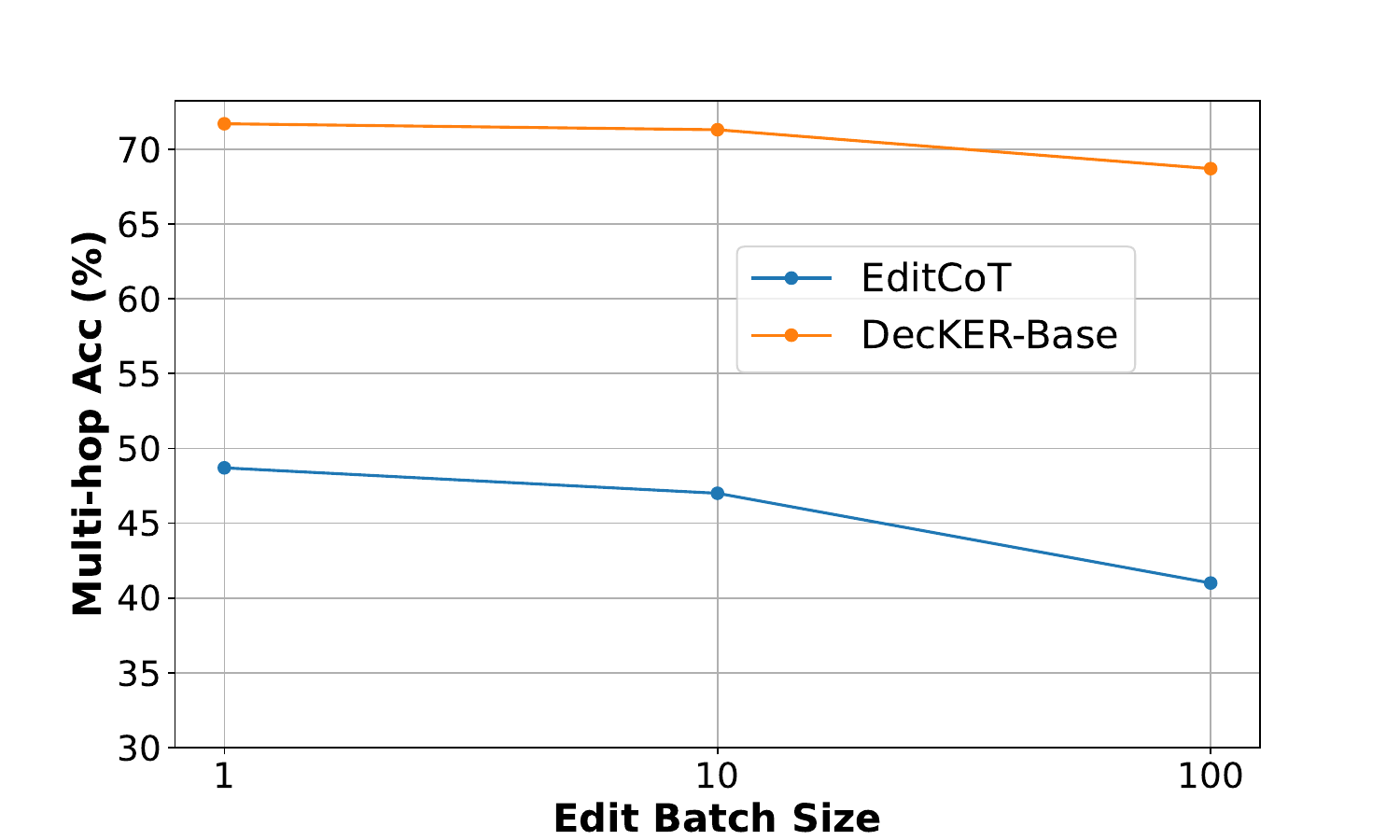}
        \caption{Edit Batch Size vs. Performance. The model is Llama-3.1-8B-Instruct and the dataset is MQuAKE-300.}
        \label{fig:editbatch}
\end{figure}

We define Edit Batch Size as the number of questions in an editing batch. We conduct experiments on MQuAKE-300 and Llama-3.1-8B-Instruct, with batch sizes of 1, 10, and 100, and compare DecKER-Base with EditCoT. The results are shown in \autoref{fig:editbatch}. We observe that both methods experience a decline in performance as the batch size increases. However, our method is more robust and exhibits lower sensitivity to editbatch compared to EditCoT.

\subsection{Impact of Sample Number on DecKER-BoN}

\autoref{tab:ab_num} illustrates the effect of the number of sampled masked paths ($N$). Performance initially improves and then levels off as $N$ increases. This indicates the effectiveness of the two-round evaluation while also suggesting that a larger $N$ is not necessarily better, as an increased number of candidates can introduce noise and impact the selection process.

\begin{table}[!t]
\centering
\caption{The relationship between sampling paths and performance in DecKER-BoN. The model is Llama-3.1-8B-Instruct and the dataset is MQuAKE-300.}
\resizebox{0.9\linewidth}{!}{
\begin{tabular}{lclll}
\toprule
\textbf{DecKER-Base}                 & \multicolumn{4}{c}{66.0}                                                                                       \\ \midrule
\multirow{2}{*}{\textbf{DecKER-BoN}} & $N=2$                    & \multicolumn{1}{c}{$N=6$} & \multicolumn{1}{c}{$N=10$} & \multicolumn{1}{c}{$N=14$} \\ \cmidrule{2-5} 
                                     & \multicolumn{1}{l}{67.7} & 70.3                      & 69.3                       & 69.7                       \\ \bottomrule
\end{tabular}
}
\label{tab:ab_num}
\end{table}

\section{Case Study}
\label{apd:casestudy}

\begin{table*}[!t]
\caption{An example of a 2-hop problem on Meta-Llama-3.1-8B-Instruct. In the "Results" column, the red sections indicate where reasoning fails. We omit the display of intermediate steps, such as retrieval scores.}
\resizebox{\textwidth}{!}{
\begin{tabular}{llc}
\toprule
\multicolumn{3}{l}{\textbf{Question: }What is the occupation of the chairperson of the Palestine Liberation Organization?} \\ \midrule
\multicolumn{3}{l}{\begin{tabular}[c]{@{}l@{}}\textbf{Related Edits:}\\ The chairperson of Palestine Liberation Organization is \uline{Mahmoud Abbas} → \uline{Regina Ip}\\ Regina Ip works in the field of \uline{politician} → \uline{basketball player}\\ \\ \textbf{Ground Truth Reasoning Path:}\\ The chairperson of Palestine Liberation Organization is Regina Ip.\\ Regina Ip works in the field of basketball player.\end{tabular}} \\ \midrule
\textbf{Method} &
  \textbf{Results} &
  \textbf{Correctness} \\ \midrule
\textbf{PokeMQA} &
  { \begin{tabular}[c]{@{}l@{}}Entity of Question: Palestine Liberation Organization, an organization.\\ Subquestion: Who is the chairperson of the Palestine Liberation Organization?\\ Generated answer: The chairperson of Palestine Liberation Organization is Regina Ip.\\ According to Generated answer, the entity of Subquestion is: Regina Ip\\ Subquestion: What is the occupation of Regina Ip?\\ Generated answer: Regina Ip works in the field of basketball player.\\ According to Generated answer, the entity of Subquestion is: basketball player\\ \textcolor{red}{However, the chairperson of Palestine Liberation Organization is actually Mahmoud Abbas}\end{tabular}} &
  \textbf{Wrong} \\ \midrule
\textbf{RAE} &
  \begin{tabular}[c]{@{}l@{}}\textbf{Retrieved Facts:} \\Palestine Liberation Organization has chairperson named as Regina Ip.\\ Regina Ip has the religion of Catholicism.\\ Catholic Church has the religion of Catholicism.\\ Catholic Church has chairperson named as Pope.\\ \\ \textbf{Answer:}\\ \textcolor{red}{there is no information about regina ip being the chairperson of the palestine liberation organization}\end{tabular} &
  \textbf{Wrong} \\ \midrule
\textbf{EditCoT} &
  \begin{tabular}[c]{@{}l@{}}\textbf{Original CoT:}\\ The chairperson of the Palestine Liberation Organization is Mahmoud Abbas.\\ Mahmoud Abbas is a public figure associated with the Palestinian National Authority.\\ The Palestinian National Authority is a government entity, and its leaders are typically involved in politics.\\ Therefore, Mahmoud Abbas, as the chairperson of the Palestine Liberation Organization, is likely a politician.\\ \\ \textbf{Edited CoT:}\\ \textcolor{red}{The chairperson of the Palestine Liberation Organization is Regina Ip.}\end{tabular} &
  \textbf{Wrong} \\ \midrule
\textbf{DecKER} &
  \begin{tabular}[c]{@{}l@{}}\textbf{Masked Reasoning Path:}\\ The chairperson of the Palestine Liberation Organization is {[}MASK 1{]}.\\ The occupation of {[}MASK 1{]} is {[}MASK ANS{]}.\\ \\ \textbf{Filled Reasoning Path:}\\ The chairperson of the Palestine Liberation Organization is Regina Ip.\\ The occupation of Regina Ip is basketball player.\end{tabular} &
  \textbf{Right} \\ \bottomrule
\end{tabular}}
\label{tab:case1}
\end{table*}

\begin{table*}[!t]
\caption{An example of a 4-hop problem on Meta-Llama-3.1-8B-Instruct. In the "Results" column, the red sections indicate where reasoning fails. We omit the display of intermediate steps, such as retrieval scores.}
\resizebox{\textwidth}{!}{
\begin{tabular}{llc}
\toprule
\multicolumn{3}{l}{\textbf{Question: }Which city is the capital of the country where the CEO of the developer of iOS 9 holds citizenship?} \\ \midrule
\multicolumn{3}{l}{\begin{tabular}[c]{@{}l@{}}\textbf{Related Edits:}\ The developer of iOS 9 is \uline{Apple Inc.} → \uline{NeXT Computer, Inc.}\\ Steve Jobs holds citizenship in the country of \uline{United States of America} → \uline{South Korea}\\ The capital of South Korea is \uline{Seoul} → \uline{Chiavari}\\ \\ \textbf{Ground Truth Reasoning Path:}\\ iOS 9 was developed by NeXT Computer, Inc.\\ The CEO of NeXT Computer, Inc. is Steve Jobs.\\ Steve Jobs is a citizen of South Korea.\\ The capital of South Korea is Chiavari.\end{tabular}} \\ \midrule
\textbf{Method} &
  \textbf{Results} &
  \textbf{Correctness} \\ \midrule
\textbf{PokeMQA} &
  \begin{tabular}[c]{@{}l@{}}Entity of Question: iOS 9, a subclass of iOS.\\ Subquestion: Which company is iOS 9 developed by?\\ Generated answer: iOS 9 was developed by NeXT Computer, Inc..\\ According to Generated answer, the entity of Subquestion is: NeXT Computer, Inc.\\ Subquestion: Who is the CEO of NeXT Computer, Inc.?\\ Generated answer: The CEO of NeXT Computer, Inc. is Steve Jobs.\\ According to Generated answer, the entity of Subquestion is: Steve Jobs\\ Subquestion: What is the country of citizenship of Steve Jobs?\\ Generated answer: Steve Jobs is a citizen of South Korea.\\ According to Generated answer, the entity of Subquestion is: South Korea\\ Subquestion: Which city is the capital of South Korea?\\ Generated answer: The capital of South Korea is Chiavari.\\ According to Generated answer, the entity of Subquestion is: Chiavari\\ \textcolor{red}{Subquestion: Is Chiavari the capital of South Korea?}\\ \textcolor{red}{Generated answer:}\end{tabular} &
  \textbf{Wrong} \\ \midrule
\textbf{RAE} &
  \begin{tabular}[c]{@{}l@{}}\textbf{Retrieved Facts:} \\iOS 9 was developed by NeXT.\\ NeXT is located in the country of Switzerland.\\ Switzerland has its capital city named as Bern.\\ Bern capital of Switzerland.\\  \\ \textbf{Answer:}\\ \textcolor{red}{there is no information about the ceo of next}\end{tabular} &
  \textbf{Wrong} \\ \midrule
\textbf{EditCoT} &
  \begin{tabular}[c]{@{}l@{}}\textbf{Original CoT:}\\ The CEO of Apple is Tim Cook.\\ Tim Cook holds citizenship in the United States.\\ The capital of the United States is Washington, D.C.\\ \\ \textbf{Edited CoT:}\\ The CEO of the developer of iOS 9 is Steve Jobs.\\ Steve Jobs is a citizen of South Korea.\\ The capital of South Korea is Chiavari.\end{tabular} &
  \textbf{Right} \\ \midrule
\textbf{DecKER} &
  \begin{tabular}[c]{@{}l@{}}\textbf{Masked Reasoning Path:}\\ The developer of iOS 9 is {[}MASK 1{]}.\\ The CEO of {[}MASK 1{]} is {[}MASK 2{]}.\\ {[}MASK 2{]} holds citizenship in the country of {[}MASK 3{]}.\\ The capital of {[}MASK 3{]} is {[}MASK ANS{]}.\\ \\ \textbf{Filled Reasoning Path:}\\ The developer of iOS 9 is NeXT Computer, Inc..\\ The CEO of NeXT Computer, Inc. is Steve Jobs.\\ Steve Jobs holds citizenship in the country of South Korea.\\ The capital of South Korea is Chiavari.\end{tabular} &
  \textbf{Right} \\ \hline
\end{tabular}}
\label{tab:case2}
\end{table*}

We showcase two examples of Meta-Llama-3.1-8B-Instruct from MQuAKE-CF-3k-v2. \autoref{tab:case1} illustrates a 2-hop question involving two edits. None of the other three baselines can solve this problem. Although PokeMQA correctly infers "basketball player," the presence of editing knowledge in the context causes the model to revise its answer. RAE fails to accurately complete the retrieval in this example. EditCoT fails to preserve the reasoning framework of the original CoT, resulting in a disruption of the reasoning process, revealing its black-box nature. Our method closely matches the ground truth, with reasoning unaffected by editing knowledge, and accurately arrives at the final answer. 

\autoref{tab:case2} is a 4-hop question involving three edits. While PokeMQA identifies the final answer "Chiavari" during the reasoning process, the model continues to decompose the next sub-question, trying to assess whether Chiavari is the capital of South Korea because it is editing knowledge. RAE's retrieval error occurs in the intermediate step, leading to an incorrect result, demonstrating its lack of robustness. EditCoT's final answer is correct, but its reasoning process does not align with the ground truth. Our method accurately completes the reasoning and filling.

\section{Licensing}

Qwen2.5-7B-Instruct and Qwen2.5-14B-Instruct is released under the Apache License 2.0. Meta-Llama-3.1-8B-Instruct is released under the LLAMA 3.1 COMMUNITY LICENSE. MEMIT, PokeMQA, and Mello are released under the MIT license. Contriever is released under the CC BY-SA 4.0 License.

The datasets MQuAKE and RippleEdits are released under the MIT license. The objective of this research is to conduct an academic investigation, which is consistent with the stipulations outlined in the associated licensing agreement.

\section{Prompt Templates}
\label{apd:prompt1}
\subsection{Prompt Templates in Analytical Experiments}

We provide GPT-4o-mini with the following system prompt:

\textit{"Extract the relation from the knowledge triples (subject, relation, object) involved in each sentence, and return a list of relations that is equal in length to the given list of sentences. You only need to provide a JSON structure that contains one key relations, whose value is the list of relations."}

In the user input section, we provide a list, with each element representing a step in a reasoning path. We utilize GPT-4o-mini's format decoding function to ensure the output is a dictionary with the key "relations" and the value being the list of extracted relations.

\subsection{Prompt Templates of DecKER}

\autoref{tab:prompt0}, \autoref{tab:prompt1}, \autoref{tab:prompt2}, and \autoref{tab:prompt3} present the prompt templates used by DecKER. We provide a 5-shot prompt template, with the blue sections in the tables corresponding to the roles in the LLM's chat template.

\begin{table*}[]
\caption{The prompt template of masked reasoning path generation.}
\centering
\resizebox{0.9\textwidth}{!}{
\begin{tabular}{|l|}
\hline
\begin{tabular}[c]{@{}l@{}}\textcolor{blue}{System:}\\ Generate a detailed chain of thought with each reasoning step labeled as [STEP]. Replace any specific knowledge, entities, dates, or locations that are not \\ in the text of question with [MASK]. Focus only on the logical steps and flow. And then include the category of each [MASK *] in order in the chain of \\ thought as [CATEGORY], to clarify the type of entity that should be filled in.\\ \\ \textcolor{blue}{User:}\  Question: What is the capital of the country where Plainfield Town Hall is located?\\ \textcolor{blue}{Assistant:}\ [STEP] Plainfield Town Hall is located in the country of [MASK 1]. \\ \
[STEP] The capital of [MASK 1] is [MASK ANS].\\ \ [CATEGORY]\\ \ [MASK 1] [country]\\ \ [MASK ANS] [city]\\ \\ \textcolor{blue}{User:}\  Question: In which country is the company that created Nissan 200SX located?\\ \textcolor{blue}{Assistant:}\ [STEP] Nissan 200SX was created by [MASK 1].\\ \ [STEP] [MASK 1] is located in the country of [MASK ANS].\\ \ [CATEGORY]\\ \ [MASK 1] [company]\\ \ [MASK ANS] [country]\\ \\ \textcolor{blue}{User:}\  Question: Which continent is the country where the director of \"My House Husband: Ikaw Na!\" was educated located in?\\ \textcolor{blue}{Assistant:}\ [STEP] The director of \"My House Husband: Ikaw Na!\" is [MASK 1].\\ \ [STEP] [MASK 1] was educated at [MASK 2].\\ \ [STEP] [MASK 2] is located in the country of [MASK 3].\\ \ [STEP] [MASK 3] is located in the continent of [MASK ANS].\\ \ [CATEGORY]\\ \ [MASK 1] [person]\\ \ [MASK 2] [school]\\ \ [MASK 3] [country]\\ \ [MASK ANS] [continent]\\ \\ \textcolor{blue}{User:}\  Question: Who is the spouse of the US president?\\ \textcolor{blue}{Assistant:}\ [STEP] The US president is [MASK 1].\ [STEP] The spouse of [MASK 1] is [MASK ANS].\\ \ [CATEGORY]\\ \ [MASK 1] [person]\\ \ [MASK ANS] [person]\\ \\ \textcolor{blue}{User:}\  Question: Who has ownership of the developer of the Chevrolet Corvette (C4)?\\ \textcolor{blue}{Assistant:}\ [STEP] The developer of Chevrolet Corvette (C4) is [MASK 1].\\ \ [STEP] [MASK 1] is owned by [MASK ANS].\\ \ [CATEGORY]\\ \ [MASK 1] [company]\\ \ [MASK ANS] [person]\end{tabular} \\ \\ \textcolor{blue}{User:}\  Question: \textit{\{A question\}} \\ \textcolor{blue}{Assistant:} \\ \hline
\end{tabular}
}
\label{tab:prompt0}
\end{table*}

\begin{table*}[]
\caption{The prompt template of conflict detection by LLMs.}
\centering
\resizebox{0.9\textwidth}{!}{
\begin{tabular}{|l|}
\hline
\begin{tabular}[c]{@{}l@{}}\textcolor{blue}{System:}\\ Determine whether the provided Fact sentence supports filling the MASKed part in the given Sentence. Only output Yes or No.\\ \\ \textcolor{blue}{User:}\  Fact: Club Nouveau is a group.\\ Sentence: Club Nouveau originated from {[}MASK 1{]}.\\ \textcolor{blue}{Assistant:}\ No\\ \\ \textcolor{blue}{User:}\  Fact: Paris is the capital of France.\\ Sentence: The capital of France is {[}MASK ANS{]}.\\ \textcolor{blue}{Assistant:}\ Yes\\ \\ \textcolor{blue}{User:}\  Fact: The Venus de Milo is located in the Louvre Museum.\\ Sentence: The Mona Lisa is located in the {[}MASK 2{]} Museum.\\ \textcolor{blue}{Assistant:}\ No\\ \\ \textcolor{blue}{User:}\  Fact: The first person to get to the South Pole is Roald Amundsen.\\ Sentence: The first person to walk on the moon was {[}MASK 3{]}.\\ \textcolor{blue}{Assistant:}\ No\\ \\ \textcolor{blue}{User:}\  Fact: The Python programming language was created by Guido van Rossum.\\ Sentence: The creator of Python is {[}MASK 1{]}.\\ \textcolor{blue}{Assistant:}\ Yes\\ \\ \textcolor{blue}{User:}\  Fact: \textit{\{A retrieved fact.\}}\\ Sentence: \textit{\{A masked step.\}}\\ \textcolor{blue}{Assistant:}\ \end{tabular} \\ \hline
\end{tabular}}
\label{tab:prompt1}
\end{table*}

\begin{table*}[]
\caption{The prompt template of entity filling.}
\resizebox{\textwidth}{!}{
\begin{tabular}{|l|}
\hline
\begin{tabular}[c]{@{}l@{}}\textcolor{blue}{System:}\\ Given the first masked entity without any analysis. The answer you give must match the given type.  \ \ \ \  \ \ \ \ \ \ \ \ \ \ \ \ \ \ \ \ \ \ \ \ \ \ \ \ \\ \\ \textcolor{blue}{User:}\  Type of the masked entity: country.\\ Sentence: Club Nouveau originated from {[}MASK 1{]}.\\ \textcolor{blue}{Assistant:}\ United States\\ \\ \textcolor{blue}{User:}\  Type of the masked entity: country.\\ Sentence: The country where the creator of the Chevrolet Corvette (C4) is located is {[}MASK ANS{]}.\\ \textcolor{blue}{Assistant:}\ United States\\ \\ \textcolor{blue}{User:}\  Type of the masked entity: place.\\ Sentence: The Mona Lisa is located in the {[}MASK 2{]} Museum.\\ \textcolor{blue}{Assistant:}\ Louvre\\ \\ \textcolor{blue}{User:}\  Type of the masked entity: person.\\ Sentence: The first person to walk on the moon was {[}MASK 3{]}.\\ \textcolor{blue}{Assistant:}\ Neil Armstrong\\ \\ \textcolor{blue}{User:}\  Type of the masked entity: person.\\ Sentence: The creator of Python is {[}MASK 1{]}.\\ \textcolor{blue}{Assistant:}\ Guido van Rossum\\ \\ \textcolor{blue}{User:}\  Type of the masked entity: \textit{\{Type of the masked entity.\}}\\ Sentence: \textit{\{A masked sentence.\}}\\ \textcolor{blue}{Assistant:}\ \end{tabular} \\ \hline
\end{tabular}}
\label{tab:prompt2}
\end{table*}

\begin{table*}[]
\caption{The prompt template of determining if entity types match, i.e. the KI Eval.}
\resizebox{\textwidth}{!}{
\begin{tabular}{|l|}
\hline
\begin{tabular}[c]{@{}l@{}}\textcolor{blue}{System:}\\ You are given an entity along with its supposed types. Your task is to determine whether the entity matches the type it has been assigned. \\ Only output Yes or No. Yes if the entity matches the type, No otherwise.\\ \\ \textcolor{blue}{User:}\  Entity: New York\\ Assigned Type: country\\ \textcolor{blue}{Assistant:}\ No\\ \\ \textcolor{blue}{User:}\  Entity: Apple Inc.\\ Assigned Type: company\\ \textcolor{blue}{Assistant:}\ Yes\\ \\ \textcolor{blue}{User:}\  Entity: Steve Jobs\\ Assigned Type: city\\ \textcolor{blue}{Assistant:}\ No\\ \\ \textcolor{blue}{User:}\  Entity: baseball\\ Assigned Type: position\\ \textcolor{blue}{Assistant:}\ No\\ \\ \textcolor{blue}{User:}\  Entity: USD\\ Assigned Type: currency\\ \textcolor{blue}{Assistant:}\ Yes\\ \\ \textcolor{blue}{User:}\  Entity: \textit{\{An entity.\}}\\ Assigned Type: \textit{\{A type.\}}\\ \textcolor{blue}{Assistant:}\ \end{tabular} \\ \hline
\end{tabular}}
\label{tab:prompt3}
\end{table*}

\end{document}